# КОМП'ЮТЕРНІ ЗАСОБИ, МЕРЕЖІ ТА СИСТЕМИ

*A. Palagin, N. Petrenko, K. Malakhov*

## TECHNIQUE FOR DESIGNING A DOMAIN ONTOLOGY

*The article describes the technique for designin a domain ontology, shows the flowchart of algorithm design and example of constructing a fragmentof the ontology of the subject area of Computer Science is considere.*
*Key words: computer ontology, ontograf, subject area, Protégé.*

*Описується методика проектування онтології ПдО, подана блок-схема алгоритму проектування ПдО і розглянуто приклад побудови фрагмента онтології з предметної області "Обчислювальна техніка".*
*Ключові слова: комп'ютерна онтологія, онтограф, предметна область, Protégé.*

*Описывается методика проектирования онтологии ПдО, приведена блок-схема алгоритма и рассмотрен пример построения фрагмента онтологии ПдО "Вычислительная техника".*
*Ключевые слова: предметная область, компьютерная онтология, онтограф, Protégé.*

УДК 004.415

А.В. ПАЛАГИН, Н.Г. ПЕТРЕНКО, К.С. МАЛАХОВ

## МЕТОДИКА ПРОЕКТИРОВАНИЯ ОНТОЛОГИИ ПРЕДМЕТНОЙ ОБЛАСТИ

**Введение**. Под компьютерной онтологией предметной области (ПдО) понимается тройка [1, 2]: $O = <X, R, F>$,

где $X = \{x_1, x_2, ..., x_i, ..., x_n\}$, $i = \overline{1, n}$,

$n = Card\ X$ – конечное множество концептов (понятий) заданной ПдО;

$R = \{r_1, r_2, ..., r_k, ..., r_m\}$, $R: x_1 \times x_2 \times ... \times x_n$,

$k = \overline{1, m}$, $m = Card\ R$, – конечное множество семантически значимых отношений между концептами ПдО. Они определяют тип взаимодействия между понятиями. В общем случае, отношения делят на общезначимые (из которых выделяют, как правило, отношения частичного порядка) и конкретные отношения заданной ПдО;

$F: X \times R$ – конечное множество функций интерпретации, заданных на концептах и/или отношениях. Частным случаем задания множества функций интерпретации $F$ является глоссарий, составленный для множества понятий $X$. Определение понятия $X_i$, в общем случае, включает подмножество понятий $\{x_{i-1}\}$, через которые определяется $X_i$; отношение $R_k$, связывающее $X_i$ с $\{x_{i-1}\}$; и множество атрибутов (признаков), присущих $X_i$.

Компьютерная онтология – формальное выражение концептуальных знаний о предметной области и по своей значимости сопоставима с базой знаний интеллектуальной информационной системы, а её построение является специфической формой человеческого мышления. Оно (мышление) в процессе познания оперирует, в том числе, суждениями, утверждениями, понятиями и отно-шениями между ними. А последние являются фундаментом, основой для построения составной части научной теории – онтологической базы знаний в заданной предметной области. При этом такие знания описываются в декларативной форме [2].

В простом случае методика проектирования онтологии ПдО (О ПдО) включает три этапа проектирования.

1. Предварительный анализ заданной ПдО.

2. Построение вручную онтографа ПдО. Под онтографом понимается двудольный граф, вершинами которого являются понятия ПдО, а дугами – связи между ними. Двудольный граф – это однонаправленный ориентированный граф, в одну вершину которого может входить и выходить несколько дуг.

3. Графическое (визуальное) проектирование онтографа ПдО и составление формализованного описания онтологии ПдО.

**Основная часть. Анализ предметной области.**

Этап системного анализа предметной области заключается в:

– составлении систематизированного представления знаний о ПдО, понимании сути происходящих в ней процессов, правил и ограничений;

– многократном абстрагировании при описании и спецификации знаний заданной ПдО, в результате которого из всего многообразия характеристик и свойств понятий предметной области выделяются наиболее существенные, релевантные множеству конкретных задач пользователей;

– составлении и документировании глоссария терминов (понятий) ПдО. В случае отсутствия в глоссарии определения для некоторого понятия (т. е. отсутствия $F$) разработчик онтологии ПдО сам определяет функцию интерпретации для этого понятия в соответствии со своими профессиональными знаниями.

Основными принципами при системном анализе ПдО, в частности, являются: принцип конечной цели; принцип единства; принцип связности; принцип иерархии; принцип развития (интеграция с другими фрагментами онтологии ПдО или «родственными» онтологиями). На этом этапе также формируются требования к разрабатываемой онтологии ПдО.

**Построение компонент онтологического описания ПдО**. Напомним некоторые известные определения, непосредственно относящиеся к построению множеств концептуальной модели ПдО или её онтологии.

Понятие – целостная совокупность суждений, в которых что-либо утверждается об отличительных признаках исследуемой сущности, ядром которой являются суждения (или утверждения) о наиболее общих и при этом существенных признаках этой сущности.

Каждое понятие характеризуется объёмом и содержанием. Объём и содержание понятия – две взаимосвязанные стороны понятия. Объём – класс обобщённых в понятии предметов, содержание – совокупность (обычно существенных) признаков, по которым произведено обобщение и выделение предметов в данном понятии.

Каждому понятию соответствует одно или несколько имён.

Все понятия (или концепты) делятся на ряд классов (по семантической зависимости) [3]:

- в зависимости от отображения вида или рода предметов – на видовые и родовые понятия;
- в зависимости от отображения части или целого предметов – на понятия-части и понятия-целые;
- в зависимости от количества отображаемых предметов – на единичные и общие понятия;
- в зависимости от отображения предмета или свойства, абстрагированного от предмета, – на конкретные понятия и абстрактные понятия.

Онтология ПдО – это концептуальная модель реального мира и её понятия должны отражать данную реальность.

**Построение фрагмента онтологии категориального уровня (ОКУ).**

Построение ОКУ для любой онтологии ПдО – важный этап в общем алгоритме проектирования:

- во-первых, обычно онтологии ПдО строятся фрагментарно. Связывание в дальнейшем нескольких таких фрагментов в общую онтологию осуществляется через понятия ОКУ;
- во-вторых, понятия О ПдО, связанные с понятиями ОКУ, являются их подклассами, и поэтому, наследуют признаки понятия-класса (конечно, если они связаны между собой отношением частичного порядка), например, отношение «род-вид» – отношение частичного порядка, а отношение «целое-часть» – нет.

Проектирование ОКУ может быть выполнено следующим образом:

- список понятий, входящих в ОКУ, может быть составлен из понятий, входящих в определения для понятий верхнего уровня О ПдО (если определение составлено согласно способу «определение понятия через понятия верхнего уровня». Понятно, что такой список будет не полным;
- пополнение списка осуществляется на основе профессиональных знаний разработчика О ПдО;
- для всех понятий, вошедших в список понятий ОКУ, составляется глоссарий (множество определений), причём, чем больше включено в глоссарий определений для одного понятия, тем точнее может быть построен онтограф ПдО и множество функций интерпретации;
- связывание дугами вершины понятий ОКУ и О ПдО. Причём, в определениях этих понятий понятия, с которыми выполняется связывание, должны быть указаны в явном виде.

Построение множества $X$ считается наиболее важным моментом при разработке онтологии ПдО. Оно должно быть обязательно не пустым.

Для хорошо проработанных предметных областей за основу множества элементов $\{x_i\}$ может быть взято содержимое различных толковых словарей. В противном случае следует составить полный список терминов, в котором указать (причём пересечение объёмов и содержаний понятий в таком предварительном списке не существенно):

– чем является каждый термин – понятием-классом предметов или конкретным понятием;

– указать для каждого термина возможные существенные отношения с другими терминами из списка;

– описать возможные существенные свойства понятий.

Следующий шаг – уточнение и определение окончательного списка классов-понятий, имена которых будут входить в разрабатываемую онтологию и являться вершинами онтографа. Также следует принять единые правила присваивания имён понятиям и свойствам, например, употребление только единственного числа, отсутствие аббревиатур и т. д. [1].

Следующим шагом является упорядочивание списка понятий по некоторому типу отношения «выше-ниже» на основе профессиональных знаний разработчика О ПдО и, возможно, следует повторить некоторые фрагменты процесса анализа ПдО (с привязкой к составленному списку понятий), выполненные на предварительном этапе.

В результате должен быть получен полный список существенных для заданной ПдО (и предполагаемых приложений) понятий и их машинно-интерпретируемые формулировки.

Построение множества $R$ также основано на результатах этапа предварительного анализа ПдО. По сути, требуется установить для каждого элемента $x_i \in X$ семантическое отношение $R_k$ с элементом $x_j \in X$, $x_i R_k x_j$, $i, j = \overline{1, n}$, $i \neq j, k = \overline{1, m}$. Другими словами, необходимо построить множество дуг, связывающих вершины направленного онтографа. В качестве вершин онтографа выступает множество понятий ПдО. Вершиной (или вершинами) онтографа (без учёта ОКУ) является родовое понятие, которое не имеет надкласса, а нижний уровень представляют конкретные понятия, т. е. не имеющие видовых понятий в заданной ПдО.

На практике множество $R$ сначала представляют некоторым обобщённым отношением "выше-ниже". Известно несколько подходов к разработке иерархии классов: процесс нисходящей разработки, процесс восходящей разработки и комбинированный процесс. Последний наиболее часто используется разработчиками, так как он является наиболее естественным, сначала оперирует понятиями среднего уровня, к которым наиболее часто обращаются разработчики. Затем эти понятия обобщаются и ограничиваются.

При связывании двух и более вершин онтографа (взятых поочередно, начиная с первых записей составленного на предыдущем этапе списка) следует извлечь информацию из соответствующих определений понятий о конкретных семантических отношениях $R_k$. В случае отсутствия такой информации или ее неполноты – отношение формируется на основе профессиональных знаний разработчика О ПдО.

В заключение данного подэтапа следует соотнести разработанные классы и их иерархии результатам предварительного анализа ПдО. В частности, уточня-ются зависимости для конкретных пар $(x_i, x_j)$. В процессе соотнесения (и построения иерархии) следует учитывать, что:

- прямые подклассы в иерархии должны располагаться на одном уровне обобщения; класс может быть подклассом нескольких классов, и тогда он может наследовать свойства от всех этих классов;
- если класс имеет только один прямой подкласс, то, возможно, при моделировании допущена ошибка или онтология неполная;
- если у данного класса есть более дюжины (иногда говорят о числе 7) подклассов, то, возможно, необходимы дополнительные промежуточные классы;
- не рекомендуется вводить в онтологию больше классов объектов, чем это необходимо для решения множества прикладных задач.

Следует помнить, что не существует единственно правильной иерархии классов.

Описанное построение онтографа является специальным видом классификации понятий ПдО – онтологической классификацией.

**Построение множества $F$.**

Для данной методики построение функций интерпретации заключается в составлении глоссария терминов ПдО, которые являются вершинами онтографа ПдО. Такой глоссарий составляется на этапе предварительного анализа ПдО, а на последующих этапах уточняется и дополняется. Причём, на этапе составления онтографа ПдО – учитывается информация (из определений понятий) о понятиях и отношениях между ними, а на этапе формирования формализованного описания – информация о существенных признаках, характеризующих определяемое понятие.

**Графическое (визуальное) проектирование онтографа ПдО и составление формализованного описания онтологии ПдО.**

На основе построенных множеств кортежа выполняется синтез концептуальной модели ПдО, например, с помощью известного инструментального средства Protégé (ИСР) и сформировать формальное описание разработанной онтологии на одном из языков описания (например, OWL), а также графическое представление онтографа.

ИСР поддерживает ручной ввод элементов множеств $X$ и $R$, в результате чего на экране получим визуальное представление онтографа ПдО. Кроме того, признаки, взятые из определений понятий, заполняются в соответствующие слоты.

В заключение в ИСР можно автоматически сформировать формализованное описание О ПдО.

Блок-схема алгоритма проектирования онтологии ПдО показана на рис. 1.

А.В. ПАЛАГИН, Н.Г. ПЕТРЕНКО, К.С. МАЛАХОВ

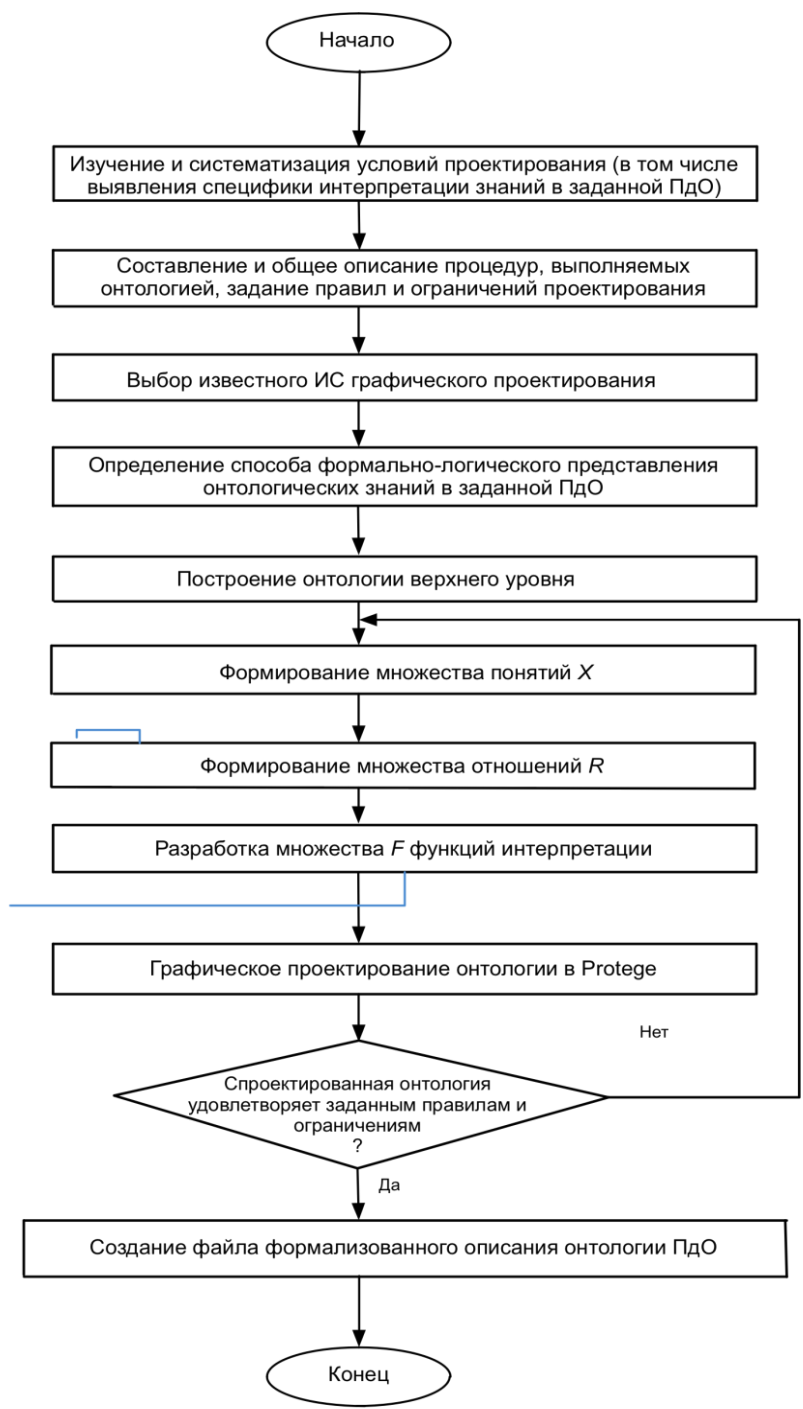

РИС. 1. Блок-схема алгоритма проектирования онтологии ПдО

Рассмотрим пример построения фрагмента онтологии из ПдО «Вычислительная техника». Из словарей по информатике и вычислительной технике выбраны следующие понятия: «Оперативная память», «Вычислительная машина», «Устройство ввода-вывода», «Аналоговая вычислительная машина», «Микропроцессор», «Центральный процессор», «Цифровая вычислительная машина», «Центральный процессор на основе микропроцессора фирмы AMD», «Информационные шины», «Управляющие шины», «Центральный процессор на основе микропроцессора фирмы Intel», «Теоретический базис», «Архитектура вычислительных систем», «Программирование», «Компьютерные сети», «Проектирование средств вычислительной техники», «Теория автоматов», «Разработчик средств вычислительной техники», «Software (программное обеспечение, ПО)», «Hardware (аппаратные средства, АС)», «Архитектура АС вычислительной системы», «Единая система стандартов», «Проектирование вычислительной системы», «АС вычислительной системы», «Архитектура вычислительной системы типа SISD», «Архитектура вычислительной системы типа MISD», «Архитектура вычислительной системы типа SIMD», «Архитектура вычислительной системы типа MIMD», «Архитектура вычислительной системы», «Программное обеспечение вычислительной системы». При этом базовыми понятиями ОКУ будут понятия «Информатика» и «Вычислительная техника».

Далее выполним ранжирование списка терминов по обобщённому отношению «выше-ниже».

1. «Информатика».
2. «Вычислительная техника», «Теоретический базис».
3. «Единая система стандартов», «Hardware», «Software», «Разработчик средств ВТ», «Архитектура вычислительных систем», «Программирование», «Компьютерные сети», «Проектирование средств вычислительной техники», «Теория автоматов».
4. «Вычислительная система (ВС)».
5. «Проектирование вычислительной системы», «АС вычислительной системы», «ПО вычислительной системы», «Архитектура ВС».
6. «Вычислительная машина», «Архитектура АС вычислительной системы».
7. «Цифровая вычислительная машина», «Аналоговая вычислительная машина», «Архитектура вычислительной системы типа SISD», «Архитектура вычислительной системы типа MISD», «Архитектура вычислительной системы типа SIMD», «Архитектура вычислительной системы типа MIMD».
8. «Центральный процессор», «Устройство ввода-вывода».
9. «Центральный процессор на основе микропроцессора фирмы AMD», «Центральный процессор на основе микропроцессора фирмы Intel».
10. «Оперативная память», «Микропроцессор».

Множество отношений состоит из элементов – {категорное_отношение, участник, множество-элемент, регламентировать, быть_характеристикой, род-вид, целое-часть, разработать, содержаться_в}.

На рис. 2 показан онтограф фрагмента ПдО «Вычислительная техника».

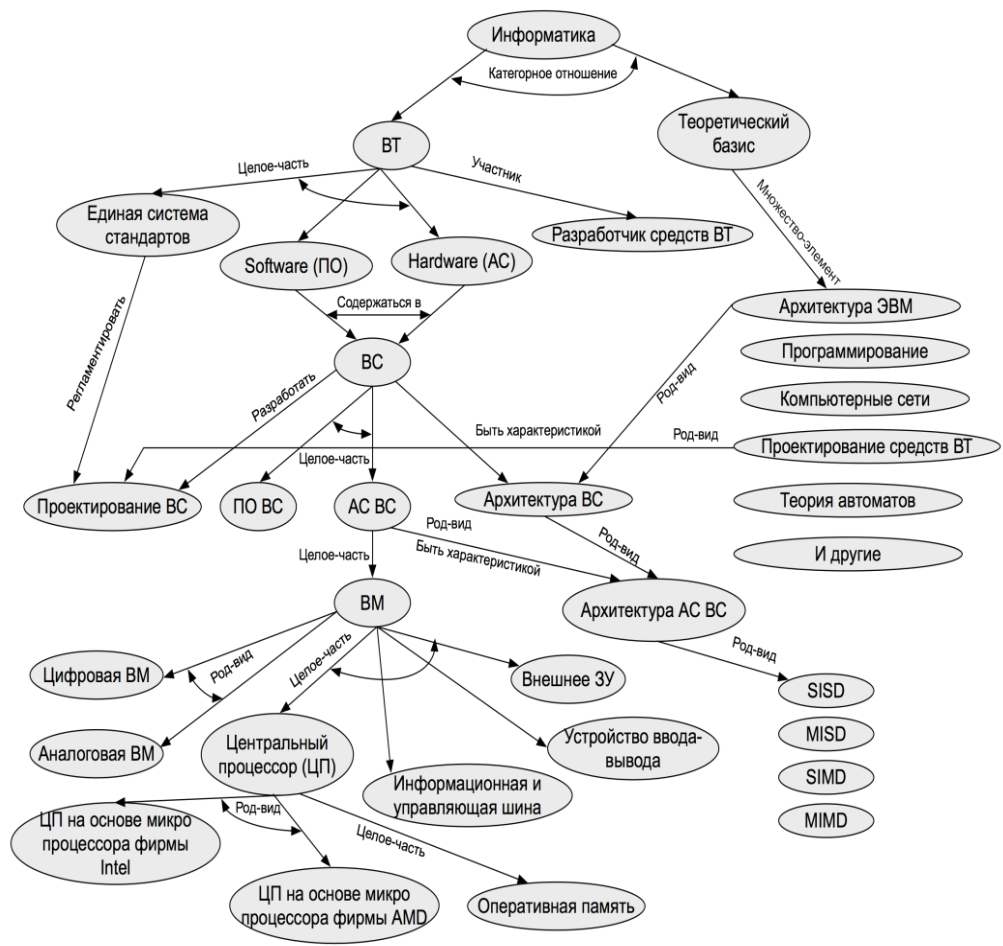

РИС. 2. Онтограф фрагмента ПдО "Вычислительная техника"

**Выводы.** Предложенная в работе методика разработки онтологии ПдО и соответствующий алгоритм ориентированы на ручное построение с автоматическим формированием их формального описания в инструментальной среде Protégé на одном из общепринятых языков описания онтологий. Для повышения эффективности процесса построения онтологий необходимо привлечение известных, или создание оригинальных инструментов автоматизированного приобретения новых знаний из различных источников.

1. *Палагин А.В., Яковлев Ю.С.* Системная интеграция средств компьютерной техники. – Винница: УНІВЕРСУМ, 2005. – 680 с.
2. *Палагин А.В., Петренко Н.Г.* Системно-онтологический анализ предметной области // УСиМ. – 2009. – № 4. – С. 3–14.
3. *Ивлев Ю.В.* Логика: учебник для вузов. – М.: «Логос», 1997. – 272 с.